\documentclass{article}
\usepackage{spconf,amsmath,epsfig, hyperref}
\usepackage{graphicx}
\usepackage{xcolor}
\usepackage{booktabs}
\usepackage{placeins}
\usepackage{comment}
\title{USING MULTI-TEMPORAL SENTINEL-1 AND SENTINEL-2 DATA\\ FOR WATER BODIES MAPPING}

\name{\begin{tabular}{c}Luigi Russo$^{a,b,1}$, Francesco Mauro$^{b}$,  Babak Memar$^{a}$, Alessandro Sebastianelli$^{c}$, \\Paolo Gamba$^{a}$
\textit{and Silvia Liberata Ullo}$^{b}$\end{tabular}\thanks{
$^{1}$Corresponding author. 
\textit{Email addresses}: luigi.russo02$@$universitadipavia.it (LR), f.mauro$@$studenti.unisannio.it (FM), babak.memar$@$uniroma1.it
 (BM), alessandro.sebastianelli$@$esa.int (AS),  paolo.gamba@unipv.it (PG), ullo$@$unisannio.it (SLU)} 
}

\address{
$^{a}$ Engineering Department, University of Pavia, Pavia, Italy \\
$^{b}$ Engineering Department, University of Sannio, Benevento, Italy \\
$^{c}$ $\phi$-lab, European Space Agency, Frascati, Italy
}

\begin{document}
\maketitle
\begin{abstract}
\noindent Climate change is intensifying extreme weather events, caus\nobreak ing both water scarcity and severe rainfall unpredictability, and posing threats to sustainable development, biodiversity, and access to water and sanitation. This paper aims to provide valuable insights 
%The proposed methodology provides valuable insights 
for comprehensive water resource monitoring under diverse meteorological conditions. An extension of the SEN2DWATER dataset is proposed to enhance its capabilities for water basin segmentation. Through the integration of temporally and spatially aligned radar information from Sentinel-1 data with the existing multispectral Sentinel-2 data, a novel multisource and multitemporal dataset 
%spanning six years 
is generated. Benchmarking the enhanced dataset involves the application of indices such as the Soil Water Index (SWI) and Normalized Difference Water Index (NDWI), along with an unsupervised Machine Learning (ML) classifier (\textit{k}-means clustering). Promising results are obtained and potential future developments and applications arising from this research are also explored.
\end{abstract}


\section{Introduction}
Climate changes are having an impact on the occurrence of extreme events, such as droughts and water scarcity on the one hand, and floods and landslides on the other hand.
\newline \noindent Extreme weather events are making water availability more scarce, more unpredictable, more polluted, or all three. These impacts throughout the water cycle threaten sustainable development, biodiversity, and people’s access to water and sanitation (\href{https://www.example.com}{Water and Climate Change}).
\newline \noindent Ensuring that everyone has access to sustainable water and sanitation services is a critical climate change mitigation strategy for the years ahead as highlighted by the Organization of United Nations (ONU) (\href{https://sdgs.un.org/goals/goal}{Sustainable Development Goal (SDG) 6: Ensure access to water and sanitation for all}).
\newline \noindent In line with this goal, we propose a research work that aims at achieving precise monitoring and mapping of water bodies and reservoirs by harnessing a multisource and multitemporal dataset spanning six years.

We propose a refinement of our SEN2DWATER dataset \cite{mauro2023, water_mit}, which is a spatiotemporal dataset generated from multispectral Sentinel-2 data gathered over water bodies ranging from July 2016 to December 2022. This refinement involves integrating the existing dataset with temporally and spatially aligned radar information from Sentinel-1 data.
\newline \noindent The result is a novel multisensor and multitemporal dataset, which, to the best of our knowledge, is unique when compared to other state-of-the-art (SOTA) datasets: (\href{https://www.kaggle.com/datasets/mateuszst/water-body-segmentation-from-satellite-images}{Water Body Segmentation From Satellite Images}, \href{https://www.kaggle.com/datasets/franciscoescobar/satellite-images-of-water-bodies}{Satellite Images of Water Bodies},  \href{https://www.kaggle.com/code/shirshmall/water-body-image-segmentation-u-net-arch/notebook}{Water Body Image Segmentation}, and works \cite{Sui2022, Feng2016, Pekel2016}).
In Table \ref{tab:combined-categorization}, the comparison between our dataset and others is presented, highlighting distinctions in features pertinent to assessing water resource dynamics.

\begin{table}[!ht]
    \centering
    \caption{Comparison with other SOTA datasets}
    \label{tab:combined-categorization}
    \resizebox{1\columnwidth}{!}{
        \begin{tabular}{llllll} 
            \toprule
            Paper/Dataset & Satellite & Multisensor & Multitemporal & Resolution & $N^\circ$ Samples \\
            \toprule
            \href{https://www.kaggle.com/datasets/mateuszst/water-body-segmentation-from-satellite-images}{Water Body Segmentation} & Sentinel-2 & No & No & 10m & 10 \\
            \href{https://www.kaggle.com/datasets/franciscoescobar/satellite-images-of-water-bodies}{Satellite Images of Water Bodies} & Sentinel-2 & No & No & 10m & 2841 \\
            \href{https://www.kaggle.com/code/shirshmall/water-body-image-segmentation-u-net-arch/notebook}{Water Body Image Segmentation} & Sentinel-2 & No & No & 10m & 2494 \\
            Sui et al.\cite{Sui2022} & Sentinel-2 & No & No & 10m & - \\
            Feng et al. \cite{Sui2022} & Landsat7 ETM+ & No & No & 30m & 8756 \\
            Pekel et al. \cite{Pekel2016} & Landsat & No & Yes & 30m & 3000000 \\
            \midrule
            \textbf{OUR DATASET} & \textbf{Sentinel-1 \& Sentinel-2} & \textbf{Yes} & \textbf{Yes} & \textbf{10m} & \textbf{12831 (S-1) \& 12831 (S-2)} \\
            \bottomrule
        \end{tabular}
    }
\end{table}

\noindent Specifically, our new dataset is a compound product of 13 optical bands of Sentinel-2 plus the 2 polarization bands (VV and VH) of Sentinel-1. Leveraging both these characteristics, our dataset enables comprehensive analysis in all atmospheric conditions with Sentinel-1 SAR data and detailed insights with Sentinel-2 high-resolution (HR) multispectral data.
\noindent
By relying on the new dataset, specific indices have been adopted for benchmarking, such as the SWI \cite{Tian2017Dynamic} for Sentinel-1 and the NDWI for Sentinel-2. Moreover, an unsupervised ML classifier has been employed, specifically the \textit{k}-means clustering algorithm, with a number of clusters equal to 4 ($\textit{k} = 4$), which was demonstrated in \cite{GambaMarzi} to be the optimal number of clusters to effectively distinguish between water bodies, vegetation, bare soil, and impervious areas. 

The research presented in this paper opens up various avenues for further exploration, such as extending the dataset and associated analyses to cover additional geographic areas and longer temporal periods. The goal is to create a comprehensive global monitoring map for water resources. In addition, incorporating ML and Deep Learning (DL) techniques can enhance the precision of water body mapping and monitoring, for advancing our understanding of climate change.

\begin{figure*}[!ht]
    \centering
    \includegraphics[width=1.7\columnwidth]{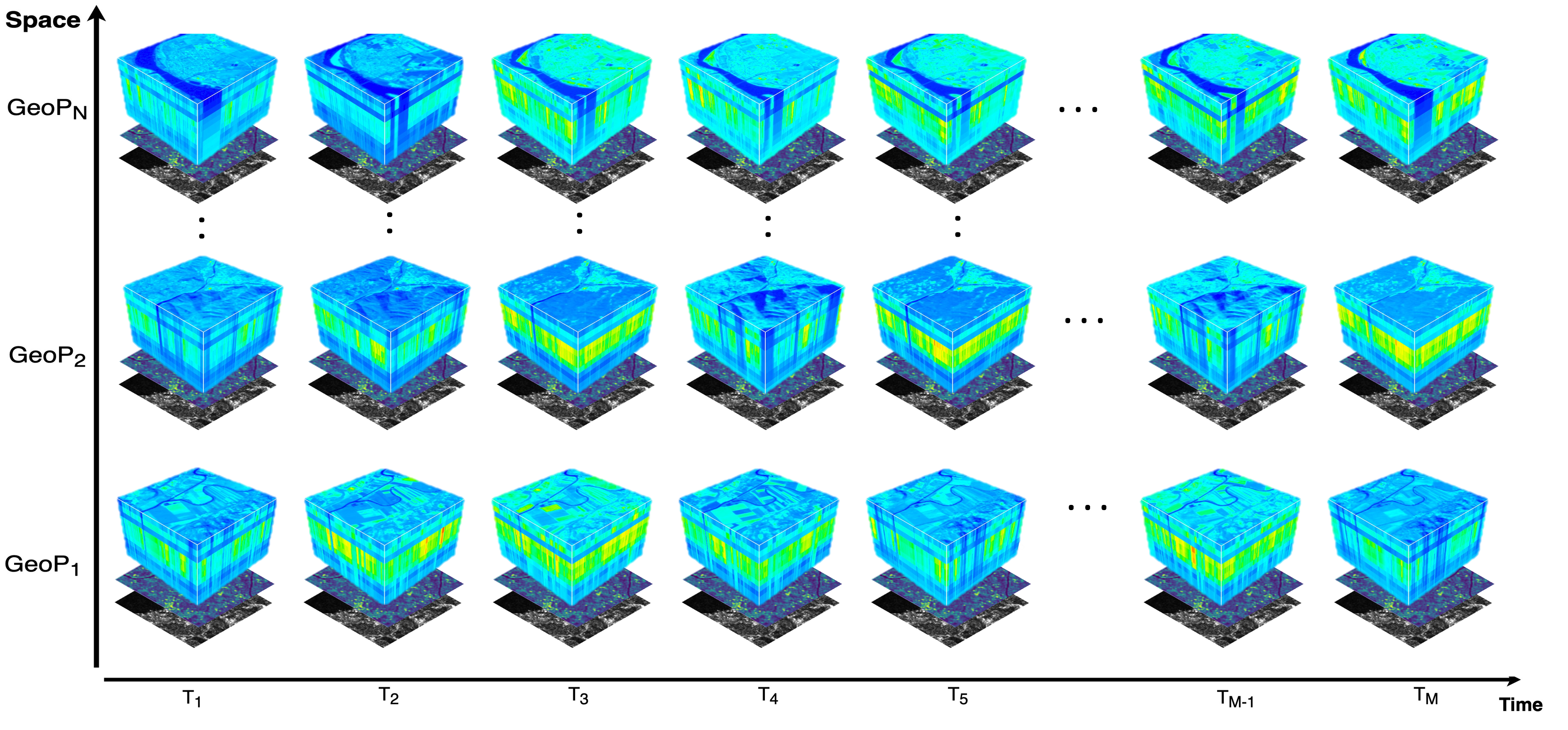}
    \caption{Visualization of the new dataset. Different geographical locations are represented on the y-axis, while the instants of each time series are represented on the x-axis for each location. Each cube plot shows the Sen1 \& Sen2 compound product composed of 13 spectral + 2 polarization bands.}
    \label{dataset}
\end{figure*}

\section{Dataset Creation}
Initially, SEN2DWATER consisted solely of Sentinel-2 (Sen2) imagery. Sen2 mission employs a wide swath width of up to 290 kilometers and a short revisit time of 5 days. This config\nobreak uration enables frequent, HR optical imaging for applications like land cover mapping and environmental monitoring.
However, the problem related to the use of only Sen2 data was related to the absence of information during the presence of clouds. Indeed, the application of this type of data products is limited by their sensitivity to weather conditions during the acquisition, as clouds can obscure portions of water bodies in images acquired by optical satellite sensors \cite{li2021mapping}.

On the other hand, the Sentinel-1 (Sen1) mission involves two satellites \footnote{
The twin Sentinel-1B satellite ended working on August 2022} designed for day and night operations, equipped with a C-band Synthetic Aperture Radar (SAR) sensor, a swath width of 250 km, and a 6-day repeat cycle. 
This configuration allows the satellites to capture radar imagery regardless of weather conditions. Anyway, it is worth highlighting that the SAR images have a non-intuitive visual appearance and this poses the biggest obstacle in SAR image annotation \cite{zhao2020opensarurban}, if compared to the accurate spatial details provided by the optical imagery.\\
%Therefore, our approach aims to overcome the limitations associated with both Sentinel-1 and Sentinel-2 data by introducing a spatiotemporal aggregation of optical and radar data. This enhancement involves integrating radar data into the existing SEN2DWATER dataset, creating a comprehensive water resource monitoring dataset. This integration ensures accurate spectral details from Sentinel-2 and all-weather monitoring capabilities from Sentinel-1, addressing the challenge of cloud cover affecting Sentinel-2 optical imagery. This combined dataset offers a more robust and versatile solution for water body monitoring under various meteorological conditions.
\noindent
Our approach aims to overcome limitations in both Sen1 and Sen2 data by enhancing the SEN2DWATER dataset, integrating radar data and forming a comprehensive water-resource-monitoring dataset,
which provides a more robust and versatile solution for mapping and monitoring water bodies under diverse meteorological conditions.

The new dataset is depicted in Fig. \ref{dataset}. It is characterized by $N$ as the number of distinct geographical points $(GeoP_n)$ and $M$ as the length for each of the time series. As specified before, the multispectral dataset SEN2DWATER for the selected water basins has been integrated with Sen1 data, ensuring spatiotemporal alignment with Sen2 and a maximum temporal difference of 5 days.
%, consistent with the revisit time of Sentinel-2. 
Specifically, the COPERNICUS/S1\_GRD\_FLOAT collection from Google Earth Engine (GEE) was employed, capturing raw power values in the Interferometric wide (IW) mode during ascending orbit passes without log scaling transformation. This resulted in the creation of a multisensor and multitemporal data aggregation spanning six years, from July 2016 to December 2022. 

\begin{figure}[!ht]
    \centering
        \includegraphics[width=0.6\columnwidth]{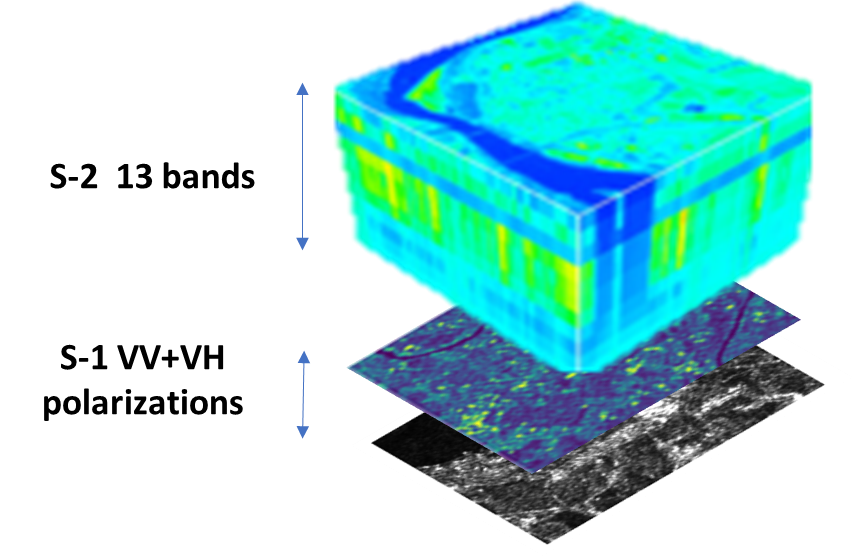} 
    \caption{Our Sentinel-1 and Sentinel-2 datacube.}
    \label{S1_S2_datacube}
\end{figure}

Each downloaded Sen1 image has been resampled to a 10 m resolution, aligning with the spatial resolution of the optical bands in Sen2 images of the old SEN2DWATER dataset. Thus, considering that the images were acquired over a polygonal area of $3 \, \text{km} \times 3 \text{km}$, each image consists of $300 \, \text{px} \times 300 \text{px}$. Therefore, our final dataset $D$ is defined by the following domain: $D \in R^{(\textit {Geo x Time } \times \textit { Width } \times \textit { Height } \times \textit { Spectrum })}$, where, in our case, $Geo=329$, $Time=39$, $Width = 300$, $Height=300$, and $\textit{Spectrum}= 15 \text{ }(13 \text{ }(\textit{Sen2}) + 2 \text{ }(\textit{Sen1}))$. This configuration defines a spatiotemporal dataset ($Geo \times Time$) comprising 12.831 Sentinel-1 images along with their corresponding 12.831 Sentinel-2 images, all spatiotemporally aligned. In Fig. \ref{S1_S2_datacube} the single datacube stacking Sen1 and Sen2 data is shown. 

Water masks from the SAR and optical images were calculated through indices such as the SWI and the NDWI to benchmark the new dataset. Additionally, this evaluation was enhanced by employing a \textit{k}-means clustering algorithm as an unsupervised ML method to accurately classify water from other Land Cover (LC) classes. To evaluate the results of this benchmarking process, we utilized a reference collection of Ground Truth (GT) data, downloaded from GEE. This collection, named \href{https://dynamicworld.app/}{Dynamic World}, is a high-resolution (10m) near-real-time (NRT) Land Use Land Cover (LULC) dataset containing class probabilities and label information for nine distinct LC categories.

\begin{comment}
    During the data download process, we specifically focused on the "water class". 
%ensuring that we work with high-resolution and near-real-time information for the "water" category within the LULC dataset. 
By leveraging %the class probabilities and 
the label information provided by this dataset, we can conduct a more detailed and accurate assessment of how well our methods performs in the classification and detection of water areas.
\end{comment}

\begin{comment}
    \begin{figure}[!ht]
    \centering
        \includegraphics[width=0.7\columnwidth]{images/datacube_s1_s2.png} 
    \caption{Our Sentinel-1 and Sentinel-2 datacube. Each image of our dataset composed of 2 radar bands (VV and VH) from Sentinel-1 plus 13 optical bands from Sentinel-2.}
    \label{S1_S2_datacube}
\end{figure}
\end{comment}

\section{Dataset Benchmarking}
Given that our dataset provides a multisensor aggregation of radar and optical data, various benchmarking applications can be explored. In particular, three different water mask techniques were investigated in this study.
\newline \noindent \textbf{The first technique} involves utilizing only Sen1 data. To extract water-related information, the formula expressed by equation (1) has been employed to calculate the SWI \cite{Tian2017Dynamic}:
\begin{equation}\label{swi_formula}
\resizebox{1\columnwidth}{!}{
$\text{SWI} = 0.1747 \times \beta_{vv} + 0.0082 \times \beta_{vh} \times \beta_{vv} + 0.0023 \times \beta_{vv}^{2} - 0.0015 \times \beta_{vh}^{2} + 0.1904$
}
\end{equation}

\noindent The terms $\beta_{vv}$ and $\beta_{vh}$ refer to the VV and VH polarizations and a threshold of 0.2 is used to distinguish water from non-water areas.
\newline \noindent \textbf{A second technique} explores a  method exclusively based on Sen2 data, as conducted in \cite{mauro2023}, and it involves utilizing the NDWI. This index is used to detect water bodies and monitor changes in water content using specific bands, typically the near-infrared (\textit{NIR}) and \textit{Green} bands.  

\begin{equation}\label{ndwi_formula}
\centering 
NDWI= \frac{\textit{Green} - NIR}{\textit{Green} + NIR}
\end{equation}

%\noindent These two approaches are illustrated in the workflow of Fig. \ref{swi_ndwi_workflow}. The two Sen1 polarizations are used to compute the SWI, and some of the Sen2 bands are employed to calculate the NDWI. The two water maps are finally compared with the GT, from which the water class is extracted, and a \textit{NaN} filtering operation is computed. \textcolor{red}{(What is NAN? va specification. Altra cosa le scritte vertical alla fine della Fig.3 non si leggono forse nel pdf ma bisogna controllare).}

\begin{figure}[!ht]
    \centering
        \includegraphics[width=1.1\columnwidth]{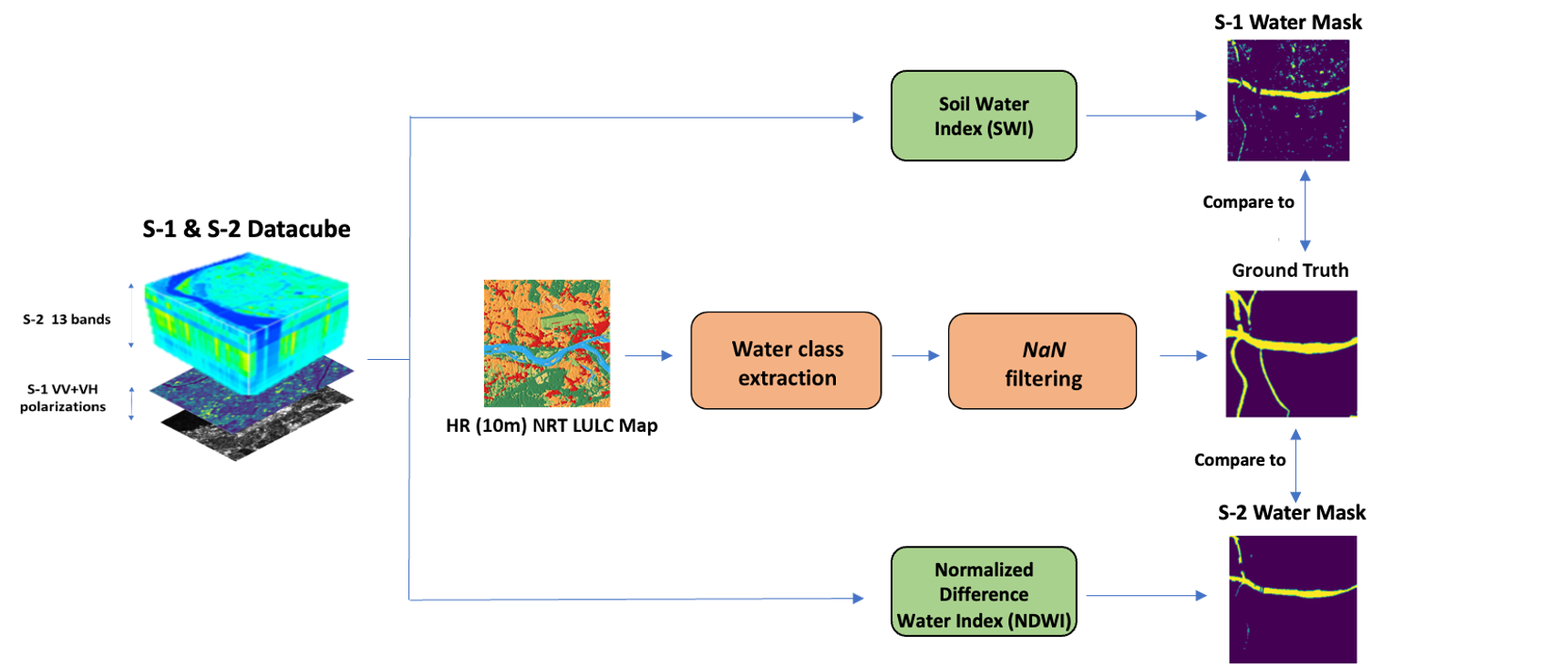} 
    %\caption{Workflow of the proposed methodology based on water class extraction from our S-1 \& S-2 datacube using the NDWI and SWI indices, respectively, and final comparison with the Ground Truth collection.}
    \caption{Workflow of the first and second methods based on the computation of the SWI and NDWI indices.}
    \label{swi_ndwi_workflow}
\end{figure}

\noindent These two approaches are illustrated in the workflow of Fig. \ref{swi_ndwi_workflow}. The two Sen1 polarizations are used to compute the SWI, and some of the Sen2 bands are employed to calculate the NDWI. The two water maps are finally compared with the GT, from which the water class is extracted, and a \textit{NaN (Not a Number)} values filtering operation is computed. This is done to ensure that all numerical values within the GT are representable, allowing value-to-value comparisons with the outcomes from the three proposed approaches.

\begin{figure}[!ht]
    \centering
        \includegraphics[width=1\columnwidth]{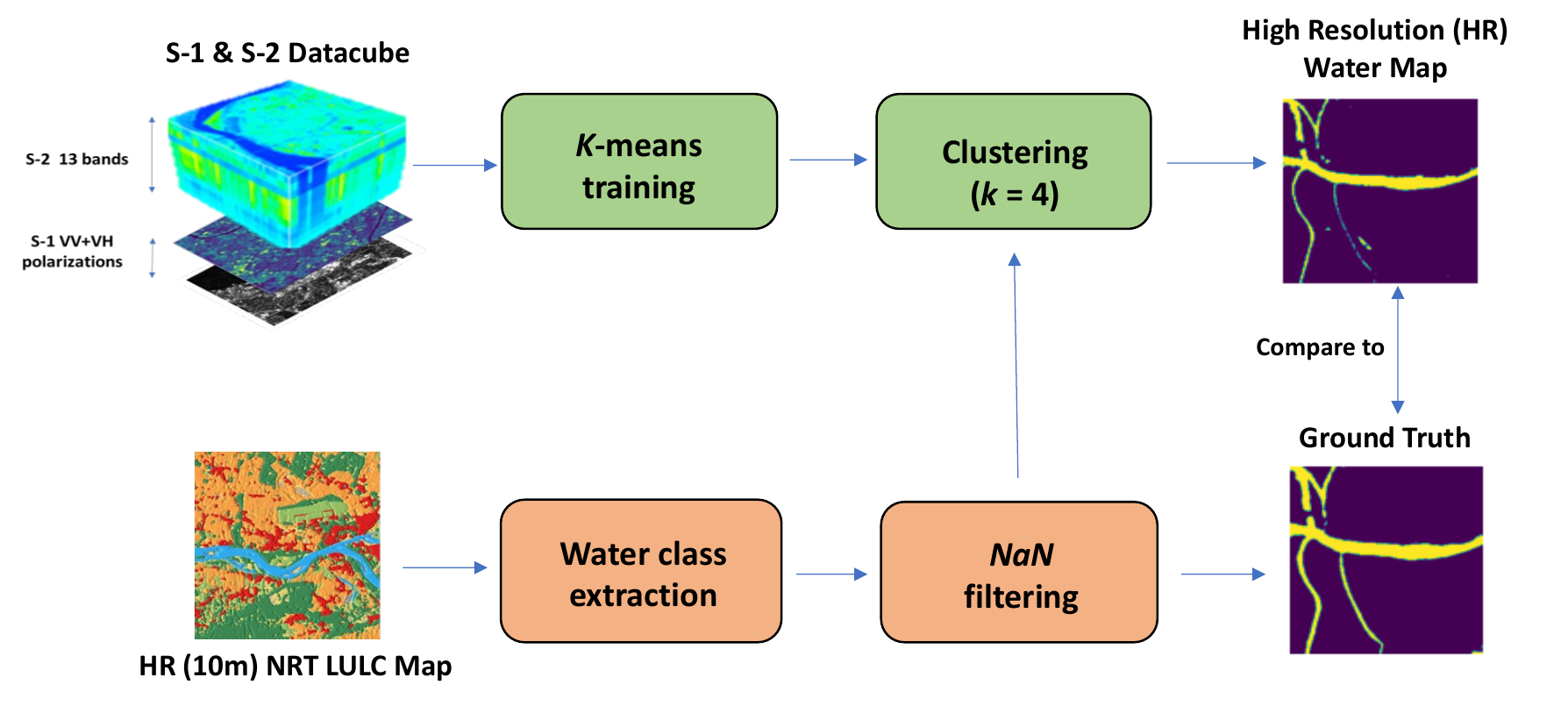} 
    %\caption{Workflow of the proposed methodology based on water class extraction from our S-1 \& S-2 datacube using the \textit{k}-means clustering algorithm and final comparison with the Ground Truth collection.}
    \caption{Workflow of the third method based on \textit{k}-means clustering algorithm.}
    \label{k_means_workflow}
\end{figure}

\noindent \textbf{The third method} employed in this study is an unsupervised ML algorithm known as \textit{k}-Means clustering. A number of clusters (\textit{k}) equal to 4 was selected, based on the work presented in \cite{GambaMarzi}, to enhance the discrimination between water and other LC classes. This classification process has utilized both Sen1 and Sen2 data from our dataset. The final results of this analysis are compared with the GT, as illustrated in Fig. \ref{k_means_workflow} where the general workflow is depicted.

%The qualitative outcomes of the three implemented methodologies are presented in Fig. \ref{visual_results}. Each row in the grid illustrates the visual results of the three proposed methods in comparison with the GT. Qualitatively, there are no significant differences among the outcomes of the three approaches.

%This qualitative assessment is further detailed in the classification report provided in Table \ref{binary_classification_metrics_table}. The table reports the percentage values for Precision, Recall, F1-Score, and Overall Accuracy (OA) metrics for the water/non-water classification task. These metrics are evaluated using a \textit{weighted average}, which computes the average of individual metrics while considering the \textit{support} (the number of true instances for each label) as weights. This operation takes into account the imbalance in the dataset by giving more importance to labels with larger support. All metrics results for the three methods are greater than or equal to 90\%, underscoring the robust performance of each approach in the water/non-water classification task.

\begin{figure}[!ht]
    \centering
        \includegraphics[width=0.8\columnwidth]{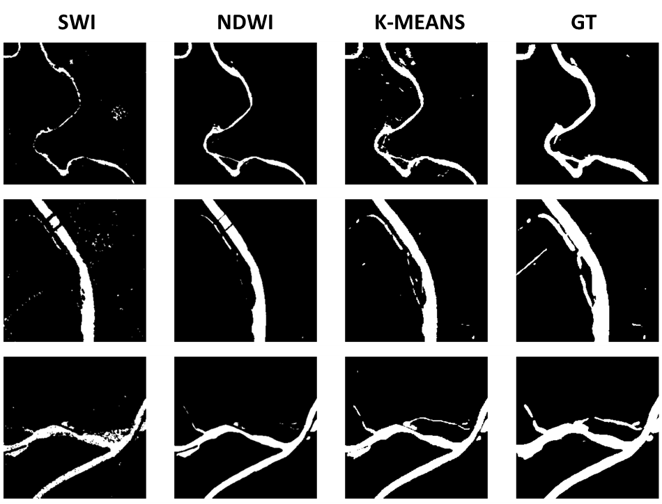} 
    \caption{Visual results for the three proposed methods in comparison with the GT.}
    \label{visual_results}
\end{figure}

The qualitative results of the three implemented methods are illustrated in Fig. \ref{visual_results}. Each row in the grid displays the visual outcomes of the three proposed methods compared to the GT. There are no significant differences among the outcomes of the three approaches, even if a higher similarity appears between the results of the third approach and the GT.

The assessment of the three proposed methods is further elaborated in the classification report presented in Table \ref{binary_classification_metrics_table}. The table includes percentage values for binary classification metrics related to the task. These metrics are computed using a \textit{weighted average}, where the \textit{support} (the number of true instances for each label) serves as weights. This approach addresses dataset imbalance by assigning more importance to labels with larger support. All metric results for the three methods are greater than or equal to 90\%, underscoring the robust performance of each approach, and showing a little better performance of the NDWI-based approach than others. Further pre-processing (i.e. despeckling) is likely to improve the outcomes, and this will be explored in future activities. Yet, the proposed methods have been deliberately applied straightforwardly, representing their advantage of an easy-to-use approach.
\begin{comment}
\begin{table}[!ht]
    \centering
    \caption{Averaged numerical results for the three methods.}
    \label{tab:mse_comparison}
    \resizebox{1\columnwidth}{!}{
        \begin{tabular}{cccc}
            \toprule
            \textbf{Score} & \textbf{SWI} & \textbf{NDWI} & \textbf{K-MEANS} \\ 
            \midrule
            \textbf{MSE $\downarrow$} & 0.092 ± 0.177 & 0.058 ± 0.133 & 0.101 ± 0.076 \\ 
            \bottomrule
        \end{tabular}
    }
\end{table}
\end{comment}

\begin{table}[!ht]
    \tiny
    \centering
    \caption{Overall accuracy (OA), precision, recall, F1-score for the three proposed methods.}
    \label{binary_classification_metrics_table}
    \resizebox{1\columnwidth}{!}{
        \begin{tabular}{cccc}
            \cmidrule[0.01pt](r){1-4}
            & \textbf{SWI} & \textbf{NDWI} & \textbf{K-MEANS} \\
            \cmidrule[0.01pt](r){1-4}
            \textbf{Precision} & 91\% & 93\% & 92\% \\
            \textbf{Recall}    & 91\% & 94\% & 90\% \\
            \textbf{F1-Score}  & 91\% & 93\% & 91\% \\
            \textbf{OA}        & 91\% & 94\% & 90\% \\           
            \cmidrule[0.01pt](r){1-4}
        \end{tabular}
    }
\end{table}

\section{CONCLUSIONS}

In this work, we introduced an innovative multisensor and multitemporal dataset, by integrating Sen1 radar data with existing multispectral Sen2 data, for water resource monitoring. The benchmarking of this dataset, using indices such as SWI and NDWI, along with the application of the \textit{k}-means clustering algorithm, demonstrated robust performance in water/non-water classification tasks.
Future developments will consider expanding the dataset to create a global and comprehensive map of water resources, including pre-processing steps, and incorporating advanced techniques such as DL,  for enhancing the methods and our understanding of climate change impacts on water availability.
%\section{REFERENCES}
\label{sec:ref}

\bibliographystyle{IEEEbib}
\bibliography{strings,refs}

\end{document}